\documentclass[letterpaper, 10 pt, conference]{template/ieeeconf}  
\IEEEoverridecommandlockouts                              
\UseRawInputEncoding 
\usepackage{graphics} 
\graphicspath{{figures/}}
\usepackage{algorithm}
\usepackage{algpseudocode}
\usepackage{stfloats}

\usepackage{amsmath} 
\usepackage{amsfonts} 
\usepackage{amssymb}  

\usepackage{xcolor}
\usepackage{colortbl}
\usepackage{array}

\usepackage[separate-uncertainty = true]{siunitx}
\usepackage[hidelinks]{hyperref}
\usepackage[capitalise]{cleveref}
\usepackage{graphicx}
\usepackage{svg}
\usepackage{bm}
\usepackage{todonotes}

\usepackage{subcaption}
\usepackage{graphicx}
\usepackage{booktabs}

\Crefname{figure}{Fig.}{Fig.} 
\Crefname{section}{Sec.}{Sec.}
\Crefname{algorithm}{Alg.}{Alg.}
\Crefname{equation}{Eq.}{Eq.}
\creflabelformat{equation}{#2\textup{#1}#3}

\usepackage[utf8]{inputenc}
\usepackage[
  backend=biber,        
  style=ieee,           
  doi=false,            
  isbn=false,           
  url=false,            
  eprint=true,          
  hyperref=true,        
  labeldateparts=false,
  maxbibnames=99,       
  sorting=none,         
  defernumbers=true,  
  maxbibnames=5,
]{biblatex}
\usepackage{xpatch}
\AtEveryBibitem{%
\ifentrytype{book}{
    \clearfield{url}%
    \clearfield{urldate}%
    \clearfield{review}%
    \clearfield{volume}%
    \clearfield{number}%
    \clearfield{pages}%
    \clearfield{note}%
    \clearfield{address}%
    \clearfield{publisher}%
    \clearfield{language}%
    \clearfield{series}
}{}
\ifentrytype{collection}{
    \clearfield{url}%
    \clearfield{urldate}%
    \clearfield{review}%
    \clearfield{volume}%
    \clearfield{number}%
    \clearfield{pages}%
    \clearfield{language}%
    \clearfield{note}%
    \clearfield{publisher}%
    \clearfield{address}%
    \clearfield{series}
}{}
\ifentrytype{incollection}{
    \clearfield{url}%
    \clearfield{urldate}%
    \clearfield{review}%
    \clearfield{volume}%
    \clearfield{number}%
    \clearfield{pages}%
    \clearfield{note}%
    \clearfield{address}%
    \clearfield{publisher}%
    \clearfield{language}%
    \clearfield{series}
}{}
\ifentrytype{article}{
    \clearfield{url}%
    \clearfield{urldate}%
    \clearfield{note}%
    \clearfield{volume}%
    \clearfield{number}%
    \clearfield{pages}%
    \clearfield{note}%
    \clearfield{address}%
    \clearfield{publisher}%
    \clearfield{language}%
    \clearfield{extra}
}{}
\ifentrytype{misc}{
    \clearfield{url}%
    \clearfield{urldate}%
    \clearfield{volume}%
    \clearfield{number}%
    \clearfield{pages}%
    \clearfield{note}%
    \clearfield{address}%
    \clearfield{publisher}%
    \clearfield{language}%
}{}
\ifentrytype{inproceedings}{
    \clearfield{url}%
    \clearfield{urldate}%
    \clearfield{note}%
    \clearfield{extra}%
    \clearfield{volume}%
    \clearfield{number}%
    \clearfield{pages}%
    \clearfield{note}%
    \clearfield{address}%
    \clearfield{language}%
    \clearfield{publisher}%
    \clearfield{eventtitle}%
}{}
}

\title{\vspace{-0.3em}\LARGE \bf
SORS: A Modular, High-Fidelity Simulator for Soft Robots
\vspace{-0.6em}
}

\author{Manuel Mekkattu, Mike Y. Michelis, and Robert K. Katzschmann$^{*}$%
\thanks{All authors are with the Soft Robotics Lab, D-MAVT, ETH Zurich, Switzerland. $^{*}$Corresponding author: \href{mailto:rkk@ethz.ch}{\tt rkk@ethz.ch}.}%
\thanks{Code and data:
\href{https://github.com/srl-ethz/sors}{\texttt{github.com/srl-ethz/sors}}.}
}

\begin{document}

\maketitle
\thispagestyle{empty}
\pagestyle{empty}

\begin{figure*}[!b]
    \vspace{-0.8em}
    \centering
    \includegraphics[width=\textwidth]{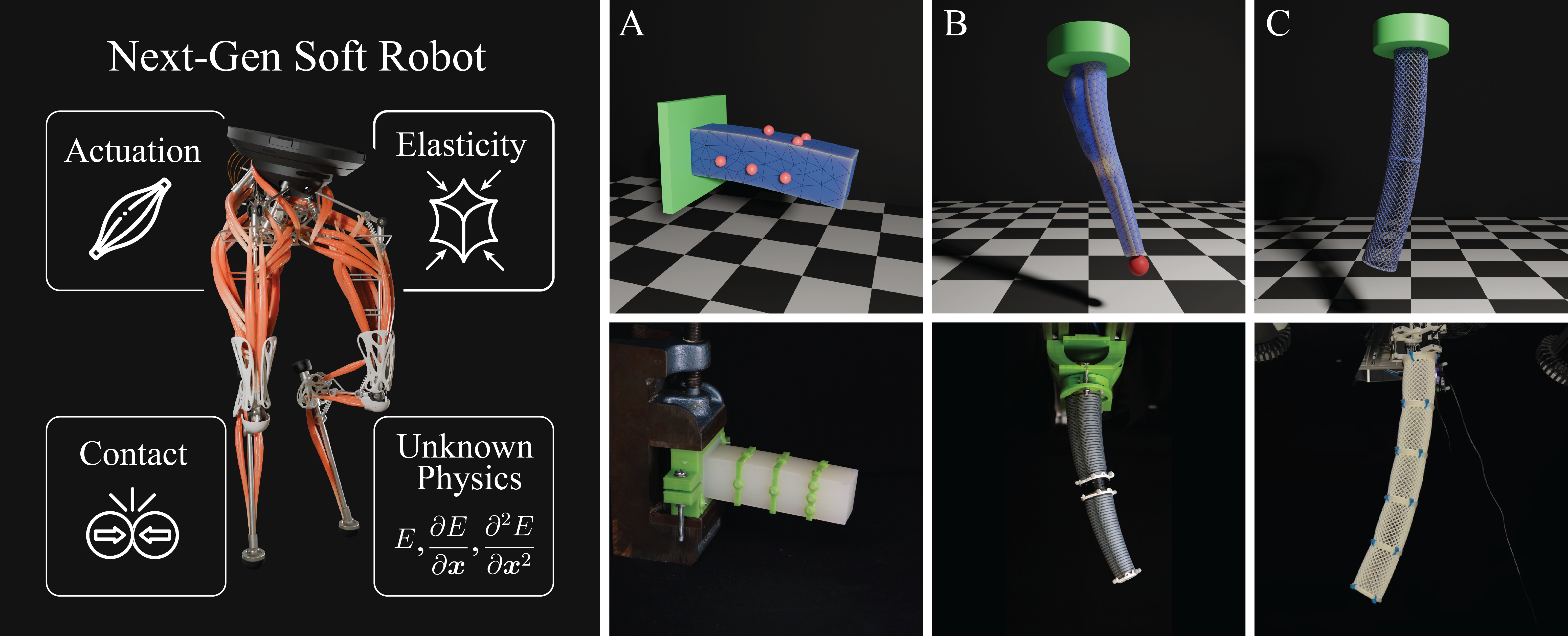}
    \caption{Next to key physical domains such as actuation, elasticity and contact, a next-gen soft robot introduces unknown physics, underscoring the need for a framework that supports customizable extensions to new physics. Our sim-to-real validated simulator addresses this challenge by enabling extensible modeling across various soft robotic systems. (A) We validate nonlinear material modeling using a cantilever as standard benchmark. (B) We reproduce real-world behavior on a pneumatic soft arm. (C) Our simulator scales up to high-resolution meshes such as tendon-driven helicoid arms (150k DoF).}
    \label{fig:overview}
    \vspace{-1em}
\end{figure*}

\begin{abstract}
The deployment of complex soft robots in multiphysics environments requires advanced simulation frameworks that not only capture interactions between different types of material, but also translate accurately to real-world performance. Soft robots pose unique modeling challenges due to their large nonlinear deformations, material incompressibility, and contact interactions, which complicate both numerical stability and physical accuracy. Despite recent progress, robotic simulators often struggle with modeling such phenomena in a scalable and application-relevant manner. We present SORS (Soft Over Rigid Simulator), a versatile, high-fidelity simulator designed to handle these complexities for soft robot applications. Our energy-based framework, built on the finite element method, allows modular extensions, enabling the inclusion of custom-designed material and actuation models. To ensure physically consistent contact handling, we integrate a constrained nonlinear optimization based on sequential quadratic programming, allowing for stable and accurate modeling of contact phenomena. We validate our simulator through a diverse set of real-world experiments, which include cantilever deflection, pressure-actuation of a soft robotic arm, and contact interactions from the PokeFlex dataset. In addition, we showcase the potential of our framework for control optimization of a soft robotic leg. These tests confirm that our simulator can capture both fundamental material behavior and complex actuation dynamics with high physical fidelity. By bridging the sim-to-real gap in these challenging domains, our approach provides a validated tool for prototyping next-generation soft robots, filling the gap of extensibility, fidelity, and usability in the soft robotic ecosystem.
\end{abstract}

\begin{keywords}
Modeling, Control, Learning for Soft Robots, Simulation and Animation.
\end{keywords}


\begin{table*}[!b]
\vspace{-0.75em}
\centering
\renewcommand{\arraystretch}{1.3}
\begin{minipage}{\textwidth}
\centering
\caption{Comparison with existing simulators.}
\label{tab:sim_comparison}
\scriptsize
\begin{tabular}{
    l
    >{\centering\arraybackslash}m{1.8cm}
    >{\centering\arraybackslash}m{3.0cm}
    >{\centering\arraybackslash}m{2.7cm}
    >{\centering\arraybackslash}m{2.4cm}
    >{\centering\arraybackslash}m{1.4cm}
    >{\centering\arraybackslash}m{1.8cm}
    >{\centering\arraybackslash}m{1.2cm}
}
\toprule
\textbf{Simulator} & \textbf{Physics Engine} & \textbf{Material Models} & \textbf{Solver Method} & \textbf{Collision Handling} & \textbf{Multimaterial} & \textbf{Extensibility}\\
\midrule 
\textbf{SORS} & FEM & Neo-Hookean, muscle material & SQP & \cellcolor{green!20}Constraint-based & \cellcolor{green!20}Yes & \cellcolor{green!20}High\\

PBD \cite{macklin_xpbd_2016} & Particle-based & Corotated linear elasticity (approx.) & Gauss–Seidel-type iteration & \cellcolor{green!20}Constraint projection & \cellcolor{red!20}No & \cellcolor{orange!20}Medium\\

SOFA \cite{faure_sofa_2012} & FEM & Linear, corotated, Neo-Hookean & Implicit and
explicit ODE solvers & \cellcolor{green!20}Constraint- and penalty-based & \cellcolor{green!20}Yes & \cellcolor{orange!20}Medium\\

OpenSim \cite{delp_opensim_2007} & Multibody dynamics & Rigid / Hill-type muscles & Implicit time integration & \cellcolor{orange!20}Spring-like contact & \cellcolor{red!20}No & \cellcolor{orange!20}Medium\\

ChainQueen \cite{hu_chainqueen_2019} & MLS-MPM & Neo-Hookean & Differentiable MPM & \cellcolor{orange!20}Grid–particle transfer & \cellcolor{green!20}Yes & \cellcolor{orange!20}Medium\\

Isaac Lab \cite{mittal_orbit_2023} & Multibody dynamics & Rigid materials & Temporal Gauss–Seidel & \cellcolor{green!20}Constraint-based & \cellcolor{red!20}No & \cellcolor{red!20}Low\\

MuJoCo \cite{todorov_mujoco_2012} & Multibody dynamics & Rigid with tendons, pneumatic and muscle actuators & Implicit velocity-stepping / complementarity-based & \cellcolor{green!20}Constraint-based & \cellcolor{red!20}No & \cellcolor{red!20}Low\\

COMSOL \cite{multiphysics_introduction_1998} & FEM & Linear, hyperelastic, viscoelastic, elastoplastic & Implicit and
explicit ODE solvers & \cellcolor{green!20}Penalty / Lagrange / augmented Lagrange & \cellcolor{green!20}Yes & \cellcolor{red!20}Low (closed)\\
\bottomrule
\end{tabular}
\end{minipage}
\vspace{-1em}
\end{table*}

\section{INTRODUCTION}
Soft robotics simulation has seen significant advances in recent years, but there remain substantial challenges in developing robust, scalable, and accurate simulation tools that can handle the complexities of real-world applications. For soft roboticists, accurate simulations are essential for designing, validating, and optimizing systems that are often resource- and time-intensive to prototype physically. These systems operate in regimes where intuition often fails: Materials deform highly nonlinearly, interactions span multiple physical domains, and actuation mechanisms are tightly coupled to morphology. Without reliable simulation tools, the design process becomes dominated by trial-and-error, limiting scalability and reproducibility. Moreover, as artificial intelligence increasingly intersects with robotics, there is a growing need for simulation frameworks that can generate large-scale, physically accurate datasets to train models for control, learning, and perception. 

Unlike rigid-body modeling, soft-body systems exhibit a high-dimensional state space due to their internal flexibility, which results in increased computational demands, more complex system dynamics, and greater sensitivity to numerical errors. 
Accurate modeling of soft materials requires constitutive laws that go beyond linear elasticity, including hyperelastic, plastic, and anisotropic formulations. 
Contact modeling is particularly difficult due to the compliant nature of the bodies involved, which leads to deformation-dependent contact areas and complex force responses.
Building a general-purpose framework that accommodates custom material models, dynamic actuation, and reliable contact resolution without compromising performance has proven technically demanding. As a result, researchers are often forced to either simplify their models or develop highly specialized code that lacks extensibility and reproducibility. This tension between generality, accuracy, and performance motivates the development of a new simulation approach tailored specifically to the needs of soft robotics.

In this work, we propose an application-driven soft body simulation framework that aims not only to enhance fidelity, but also to address a core scientific question: \textit{How can soft robot simulations be reformulated around a set of minimal, physically interpretable interfaces that generalize across materials, actuation mechanisms, and contact scenarios?} 

Based on the Finite Element Method (FEM), our framework is designed to modularize the critical components of the soft robot simulation pipeline. We introduce three core interfaces --- \textit{energies, forces}, and \textit{constraints} --- which serve as modular building blocks for defining the physical behavior of the system. Energies capture governing physics such as elasticity or damping. Forces encode actuation inputs, such as pressure or tendons. Lastly, constraints define contact conditions and boundary interactions. 
They enable researchers to plug in new actuation models or materials with minimal effort.
By releasing this platform as an open source tool, we aim to accelerate research in soft robotics by providing a principled, extensible, and validated simulation environment. We take a step toward closing the sim-to-real gap for soft robots where simulation drives the creation of safer, more adaptable, and more intelligent machines.

\section{RELATED WORK}

State-of-the-art robotics simulators often revolve around rigid robots~\cite{todorov_mujoco_2012, mittal_orbit_2023}, such as quadrupeds and robot arms, and less about deformable bodies. Rigid-body frameworks are well developed and parallelize nicely on Graphical Processing Units (GPU). On the other hand, soft object simulations remain expensive or inaccurate for real-world robotics~\cite{gao_sim--real_2024}.

Modeling soft bodies introduces fundamentally different challenges. Their deformation obeys Newton’s second law, where the internal elastic forces are defined by the chosen material model. This balance of governing forces can be reformulated as an energy optimization problem~\cite{bouaziz_projective_2014, ferguson_intersection-free_2021}, which allows the application of efficient nonlinear optimization methods. Recent methods simplify optimization by approximating the internal energies with quadratic functions~\cite{bouaziz_projective_2014}, enabling real-time simulation at the cost of physical accuracy. 
Although some methods can be coupled with rigid bodies~\cite{faure_sofa_2012, ferguson_intersection-free_2021}, a general modular approach to coupling arbitrary multiphysics is not considered in current state-of-the-art simulation. 

A widely used, open-source framework in the soft robotics community is SOFA~\cite{faure_sofa_2012}, though originally designed for computational medical simulations. SOFA has been applied to pneumatic soft robots~\cite{katzschmann_dynamically_2019, cangan_model-based_2022}, but usability and extensibility can be lacking for new applications such as biological muscle models for biohybrid robotics. Compared with the reinforcement learning community, using frameworks such as MuJoCo~\cite{todorov_mujoco_2012} or Isaac Lab~\cite{mittal_orbit_2023}, the current deformable body simulation environments are simply not as user-friendly --- they often lack \texttt{Python} bindings or easy installation, making them cumbersome to integrate into machine learning or optimization pipelines.

Existing soft-body simulators make it difficult to add new materials, actuators, or rigid–soft couplings, often requiring invasive changes to the core \texttt{C++} code. They lack a unified interface for extending physics modules, such as materials, constraints, or actuation models. Our framework overcomes this by introducing modular interfaces for energies, forces, and constraints, enabling flexible and consistent integration of new physical behaviors without altering the core system.

To contextualize SORS within existing simulation ecosystems, we provide a comparative summary of key frameworks in \Cref{tab:sim_comparison}. The table highlights the distinguishing features of our simulator relative to established rigid- and soft-body simulators, including their underlying physics engines, material modeling capabilities, collision handling strategies, and extensibility. This overview emphasizes how we close the gap between high-fidelity soft body modeling and the modularity typically only found in rigid-body simulators.

\section{METHODS}
\subsection{Energy Minimization}
The core of our simulation framework is an energy-based approach, which forms the foundation for modeling physical interactions in soft robotics and other complex systems. This approach relies on the principle that a physical system moves towards its energetic minimum. By defining energy functions that describe physical phenomena such as deformations or gravity, the system can compute how objects will behave. We formulate the problem setup as a constrained minimization problem of the total energy. Let $E(\bm{x})$ be the system energy function of variable $\bm{x} \in \mathbb{R}^n$. We define equality constraints \( f_i(\bm{x}) = 0 \), indexed by \( i \in \mathcal{E} = \{1 ... p\} \), and inequality constraints \( h_j(\bm{x}) \geq 0 \), indexed by \( j \in \mathcal{I} = \{1 ... m\} \), where \( p \) and \( m \) are the number of equality and inequality constraints, respectively. We want to solve the minimization problem
\begin{align}
\label{eq:optimization}
\min_{\bm{x}} \quad & E(\bm{x}) \\
\text{subject to} \quad & \bm{f}(\bm{x}) = \bm{0} \\
                        & \bm{h}(\bm{x}) \geq \bm{0}
\end{align}
where $\bm{f}(\bm{x}) = (f_i(\bm{x}))_{i \in \mathcal{E}}$ and $\bm{h}(\bm{x}) = (h_j(\bm{x}))_{j \in \mathcal{I}}$ are the stacked constraint vectors. We solve this using the Sequential Quadratic Programming (SQP) method, as it is particularly effective for problems with significant nonlinearities in its constraints, which is a common feature in soft robotic systems due to contact, actuation, and material behavior~\cite{jorge_nocedal_numerical_nodate}. At each SQP iteration, the problem is locally approximated by a Quadratic Program (QP), offering fast local convergence and robust handling of both equality and inequality constraints:
\begin{align}
\min_{\Delta\bm{x}} \quad & \frac{1}{2} \Delta\bm{x}^\top H\, \Delta\bm{x} + \bm{g}^\top \Delta\bm{x} \\
\text{subject to} \quad & J_f\, \Delta\bm{x} + \bm{f}(\bm{x}) = \bm{0} \\
                        & J_h\, \Delta\bm{x} + \bm{h}(\bm{x}) \geq \bm{0}
\end{align}
where $H = \nabla^2 E(\bm{x}) \in \mathbb{R}^{n \times n}$ is the Hessian of the energy function, $\bm{g} = \nabla E(\bm{x}) \in \mathbb{R}^n$ the gradient, and \( J_f \in \mathbb{R}^{p \times n} \), \( J_h \in \mathbb{R}^{m \times n} \) are the Jacobians of the equality and inequality constraints, respectively. If no constraints are present (i.e., $\mathcal{E}$ and $\mathcal{I}$ are empty), the problem reduces to an unconstrained energy minimization. In this case, the SQP iteration simplifies to a standard Newton step
\begin{align}
H\, \Delta\bm{x} = -\bm{g}.
\end{align}
The update is applied iteratively as \( \bm{x}_{k+1} = \bm{x}_k + \Delta\bm{x} \) until convergence. The modularity of our framework allows users to specify custom energy terms based on the physical properties they wish to model, such as elasticity, plasticity, or damping. Each energy function then contributes to the overall system behavior, with gradients and Hessians derived from these functions to describe forces and stiffness matrices, respectively. This allows for accurate modeling of soft materials, including their nonlinear deformations, while also supporting user-defined extensions for various physical laws. 

\subsection{Framework Architecture}

\begin{figure}[!b]
    \vspace{-1em}
    \centering
    \includegraphics[width=\linewidth]{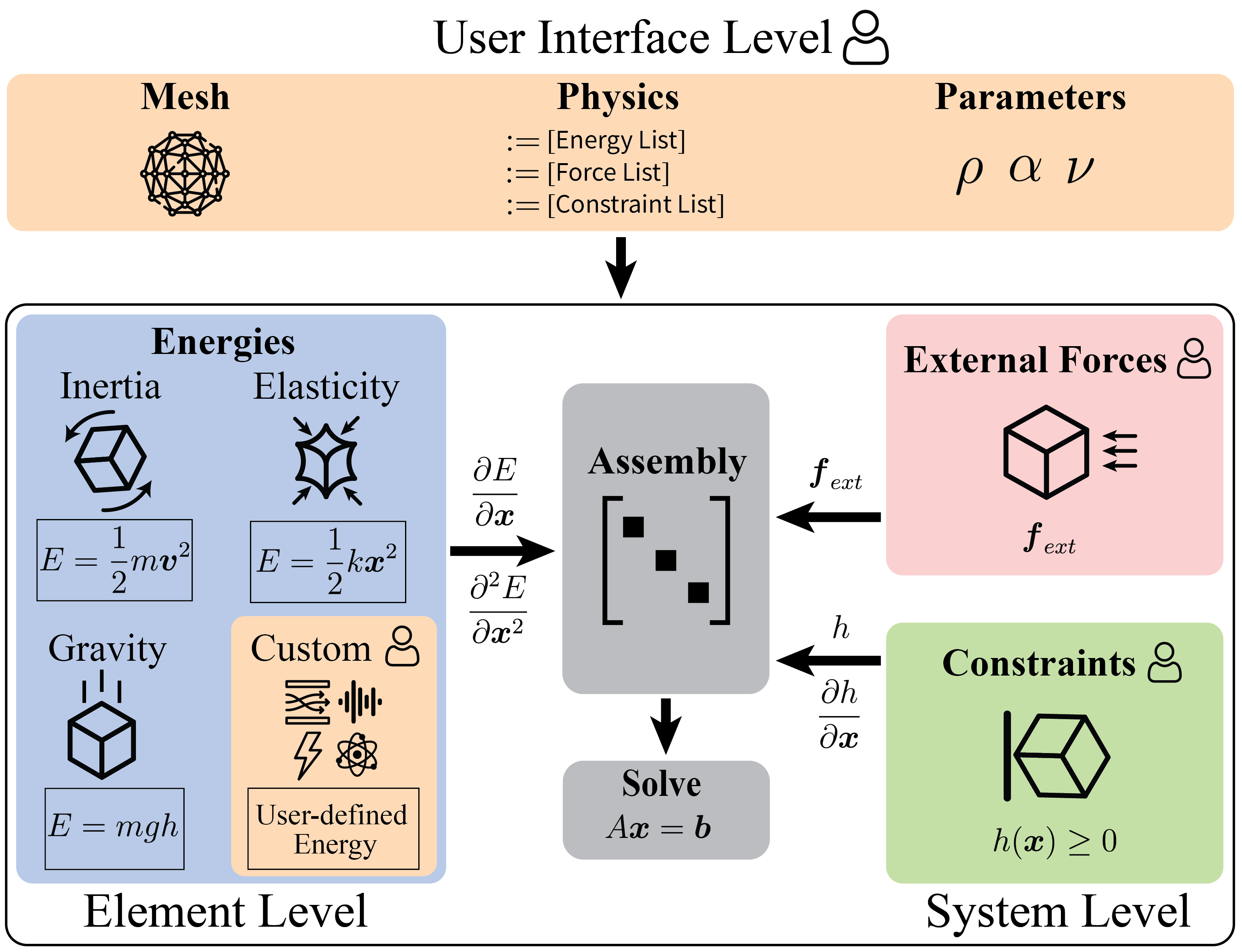}
    \caption{Framework architecture of our simulator. On element level, we compute energy terms such as inertia, gravity, and elasticity. On system level, external forces and constraints are evaluated. The user interfaces with the framework by providing custom geometry, physical laws and parameters.}
    \label{fig:architecture}
\end{figure}

Our framework is structured around three modular interfaces --- energies, forces, and constraints --- which together define the full physical behavior of a soft robotic system. Energies capture the underlying physics, such as elasticity, damping, or potential fields. Forces represent active components such as pressure. Constraints handle boundary conditions and contact interactions. The architecture distinguishes between element-level and system-level computations. Energy terms are locally evaluated at quadrature points per element, either tetrahedral or hexahedral, enabling the computation of small, independent energy gradients and Hessians. These local contributions are then assembled into global matrices governing the system state evolution. In contrast, forces and constraints operate at the system level, ensuring consistent coupling across the mesh. This locality-based decomposition not only reflects the physical structure of the problem but also provides a clear path toward future GPU parallelization, as independent element-wise energy evaluations can be computed concurrently. A schematic overview of this modular structure is shown in \Cref{fig:architecture}. 

Our framework is implemented in \texttt{C++}, using the \texttt{Eigen} library for linear algebra, and the \texttt{OSQP} solver~\cite{stellato_osqp_2020} for solving the QP subproblems. 
A \texttt{Python} front-end provides an accessible interface for simulation control, optimization, and data analysis. The modular design of the codebase allows users to define custom energy terms and constraints, which are automatically integrated into the computation pipeline. We summarize our constraint-based optimization framework in \Cref{alg:sqp}.

\begin{algorithm}
\caption{SQP for Constrained Energy Minimization}
\label{alg:sqp}
\begin{algorithmic}
\State \textbf{Choose} initial guess $\bm{x}_0$, \textbf{Set} tolerance $\epsilon$, step size $\alpha$
\For{$k = 0, 1, 2, \dots$}
    \State Evaluate gradient $\bm{g}_k = \nabla E(\bm{x}_k)$
    \State Evaluate Hessian $H_k = \nabla^2 E(\bm{x}_k)$
    \State Evaluate constraint values $\bm{f}_k = \bm{f}(\bm{x}_k)$, $\bm{h}_k = \bm{h}(\bm{x}_k)$
    \State Evaluate constraint Jacobians $J_f^k = \nabla \bm{f}(\bm{x}_k)$\\
    \hspace{4.75cm} $J_h^k = \nabla \bm{h}(\bm{x}_k)$
    \State \textbf{Solve} the Quadratic Program:
    \begin{align*}
    \min_{\Delta\bm{x}} \quad & \frac{1}{2} \Delta\bm{x}^\top H_k \Delta\bm{x} + \bm{g}_k^\top \Delta\bm{x} \\
    \text{subject to} \quad & J_f^k \Delta\bm{x} + \bm{f}_k = 0, \\
                            & J_h^k \Delta\bm{x} + \bm{h}_k \geq 0
    \end{align*}
    \State \textbf{Update}  $\bm{x}_{k+1} = \bm{x}_k + \alpha \Delta\bm{x}$
    \If{$\|\Delta\bm{x}_k\| < \epsilon$}
        \State \textbf{Return} $\bm{x}_k$ as solution
    \EndIf
\EndFor
\end{algorithmic}
\end{algorithm}

\subsection{Time Integration}

Our framework incorporates time integration via the kinetic energy term in its energy minimization approach, supporting Backward Euler and Crank-Nicolson schemes. Backward Euler ensures stability but introduces artificial dampening, while Crank-Nicolson preserves energy without numerical artifacts. An adaptive time-stepping method, guided by the Courant-Friedrichs-Lewy condition, dynamically adjusts step sizes to balance stability and computational efficiency, maintaining physical accuracy. To determine the kinetic energy of a single element, we treat each vertex as a distinct mass point and calculate the kinetic energy for each vertex separately. We note that we cannot simply calculate the kinetic energy of the element center of mass, as this would neglect inertia from rotational movement. We introduce time integration through the equations of motion 
\begin{align}
\begin{split}
    m \dot{\bm{v}}_t &= \bm{f}_{int} \left(\bm{x}_t \right)
    \\
    \dot{\bm{x}}_t &= \bm{v}_t
\end{split}
\label{eq:kinetic}
\end{align}
defined by single mass point $m$ with position and velocity $\bm{x}, \bm{v} \in \mathbb{R}^n$, and internal forces $\bm{f}_{int}$ defined by other energies such as elastic, gravitational, etc. We implement backward Euler as the following discretization of \Cref{eq:kinetic}:
\begin{align}
\begin{split}
    m \dfrac{\bm{v}_t - \bm{v}_{t-1}}{\Delta t} &= \bm{f}_{int} \left(\bm{x}_t \right) \\
    \dfrac{\bm{x}_t - \bm{x}_{t-1}}{\Delta t} &= \bm{v}_t
\end{split}
\label{eq:backwardEuler}
\end{align}
where we need to keep track of previous positions and velocities to solve for the current timestep $t$. The final variable we solve for is $\bm{x}_t$, which computes the values for $\bm{v}_t$ and $\bm{f}_{int} \left(\bm{x}_t \right)$. Similarly, we extend this to the energy-conserving Crank-Nicolson integration scheme
\begin{align}
\begin{split}
    m \dfrac{\bm{v}_t - \bm{v}_{t-1}}{\Delta t} &= \frac{1}{2} \left[ \bm{f}_{int} \left(\bm{x}_t \right) + \bm{f}_{int} \left(\bm{x}_{t-1} \right) \right] \\
    \dfrac{\bm{x}_t - \bm{x}_{t-1}}{\Delta t} &= \frac{1}{2} \left[ \bm{v}_t + \bm{v}_{t-1} \right]
\end{split}
\label{eq:CN}
\end{align}
where we now also need to store the previous internal forces. These time integrators can be easily extended, and enable consistent coupling between temporal derivatives and energy-based dynamics across integration schemes.

\subsection{Neo-Hookean Elastic Energy}

The Neo-Hookean model, a nonlinear extension of Hooke's law, is widely used to characterize elastic materials under deformation~\cite{smith_stable_2018}. It can model the large-strain behavior of soft robotic materials, and forms a cornerstone of the framework. Numerous variations of the Neo-Hookean model have been developed to address different material behaviors~\cite{ogden_non-linear_1997, bower_applied_2009}. In this work, we adopt the most commonly used formulation of Neo-Hookean elasticity~\cite{bonet_nonlinear_1997}, where the energy function for a single element is defined as
\begin{align}
\label{eq:neohook}
E_{\text{NH}} = \left( \dfrac{\mu}{2} \left( \text{tr} \left[ F^\top F \right] - 3 \right) - \mu \ln{J} + \dfrac{\lambda}{2} \left(\ln{J}\right)^2 \right) V
\end{align}
where $F$ is the deformation gradient of the element, $\text{tr}[\cdot]$ the trace, $J = \text{det} \left( F \right)$ and $V$ the element volume. $\lambda$ and $\mu$ are the first and second Lam\'e parameters, and can be expressed directly via the Poisson's ratio $\nu$ and Young's modulus $E$, two material-specific parameters. 
The deviatoric term, $\frac{\mu}{2} \left( \text{tr} \left[ F^\top F \right] - 3 \right)$, describes the shape-changing response of the material, independent of volume changes. 
The terms $- \mu \ln{J}$ and $\frac{\lambda}{2} \left(\ln{J}\right)^2$ describe volumetric terms, which penalize changes in volume.

We require the gradient and Hessian of the energy to solve \Cref{eq:optimization}, this first- and second-order information ensures accurate and stable numerical optimization during simulation. 
The gradient of the Neo-Hookean energy can be expressed component-wise, using Einstein notation, through the application of the chain rule:
\begin{align}
    \label{eq:nh_grad_comp}
    \dfrac{\partial E_{\text{NH}}}{\partial x_i} = 
    \dfrac{\partial E_{\text{NH}}}{\partial F_k} \dfrac{\partial F_k}{\partial x_i} \; V.
\end{align}
For the sake of brevity, we derive the gradient and Hessian for a tetrahedral element $\mathcal{T}$, but the derivation follows equivalently for hexahedral elements. The tetrahedron $\mathcal{T}$ has deformed vertices described by $\tilde{\bm{x}}_k \in \mathbb{R}^3$ with $k= 1 ... 4$, or after flattening the vectors $x_i \in \mathbb{R}$ with $i= 1 ... 12$. Using the deformed and undeformed vertices, the deformation gradient $F \in \mathbb{R}^{3 \times 3}$ is computed. After flattening $F$ into a vector, the index $k = 1...9$ corresponds to its 9 components. ${\partial F_k}/{\partial x_i}$ is referred to as the \textit{deformation Hessian}, representing a $9 \times 12$ tensor that encodes the derivatives of the deformation gradient components. The detailed computation of the deformation gradient and deformation Hessian can be found in existing literature, we refer the reader to~\cite{sifakis_fem_2012}. We use the matrix calculus identities~\cite{kaare_brandt_petersen_matrix_2012} for $\partial/\partial A_{ab} \left( \text{tr}[A^\top A]\right) = 2 A_{ab}$ and $\partial/\partial A_{ab} \left( \ln{\text{det($A$)}} \right) = A^{-\top}_{ab}$ with $A \in \mathbb{R}^{n \times n}$ to compute the first factor of \Cref{eq:nh_grad_comp}. Note that we obtain row-major flattening index $k = 3(a-1) + b$ from row-index $a = 1,2,3$ and column-index $b = 1,2,3$. It follows directly that the gradient of the Neo-Hookean energy is given by
\begin{align}
    \dfrac{\partial E_{\text{NH}}}{\partial x_i} = \left( \mu F_k  + \left( \lambda \ln{J} - \mu \right) F^{-\top}_k \right) \dfrac{\partial F_k}{\partial x_i} \; V.
\end{align} 
For the energy Hessian $H_{ij} = \left(\partial_i \partial_j E_{\text{NH}} \right)_{ij}$ with $i,j = 1...12$ 
\begin{align}
    H_{ij}
    &= \dfrac{\partial}{\partial F_l} \left( \left( \mu F_k + \left( \lambda \ln{J} - \mu \right) F^{-\top}_k \right) \dfrac{\partial F_k}{\partial x_i} \right) \dfrac{\partial F_l}{x_j} \; V \nonumber \\[0.5em]
    &= (\mu \delta_{kl} + \lambda F^{-\top}_l F^{-\top}_k + \left(\lambda \ln{J} - \mu\right) \nonumber \\
    &\quad \left(- F^{-\top} E_l^\top F^{-\top} \right)_k) \dfrac{\partial F_k}{\partial x_i} \dfrac{\partial F_l}{x_j} \; V
\end{align}
we use the chain rule and the matrix calculus identity $(\partial A^{-\top}/\partial A_{ab})_{ij} = -(A^{-\top} E^{\top}_{ab} A^{-\top})_{ij}$ for $A \in \mathbb{R}^{n \times n}$ and $(E_{ab})_{ij} = \delta_{ai} \delta_{bj}$. 
By following a similar procedure, the gradient and Hessian can be computed for a broad range of material models, providing a versatile framework for diverse applications. This, for example, includes a more stable formulation of the Neo-Hookean energy~\cite{smith_stable_2018} that can handle mesh inversions which is used in \Cref{sec:poke}.

\subsection{Pressure Actuation}

Many soft robots are actuated through internal pressure, which generates motion and deformation by exerting normal forces on internal surfaces. 
We incorporate pressure actuation as an external force added to the total energy gradient. Consider an internal surface triangle with vertex positions \( \bm{x}_0, \bm{x}_1, \bm{x}_2 \in \mathbb{R}^3 \). The corresponding surface normal is computed as $\bm{n} = \frac{1}{2} \left( \bm{x}_1 - \bm{x}_0 \right) \times \left( \bm{x}_2 - \bm{x}_0 \right)$
where \(\bm{n}\) encodes both orientation, and area via $\|\bm{n}\|$. Applying a uniform internal pressure $p$ to this surface results in a force acting opposite to the outward normal:
\begin{align}
\bm{f}_p = -p \bm{n}.
\end{align}
This force is distributed equally among the element vertices, producing a contribution of \( \bm{f}_p / 3 \) at each node. 
If multiple triangles are pressurized, their contributions are accumulated across all affected surface vertices. 

\subsection{Muscle Actuation}

We follow previous work \cite{min_softcon_2019} to implement active control into our finite elements. We define the energy as
\begin{align}
    E_{m} &= \frac{k}{2} \|(1-a) F \bm{m}\|^2 \; V
\end{align}
with $F$ the deformation gradient as a function of $\bm{x}$, $k$ the muscle stiffness, $a$ the actuation strength which can change each timestep, and $\bm{m} \in \mathbb{R}^3$ the actuation direction. The actuation $a$ is one-dimensional, along a given direction $\bm{m}$, where a value of $a=0$ means no contraction, positive values $a>0$ contraction, and negative $a<0$ expansion.

\subsection{Mass-Proportional Damping}

Since we have an energy-conserving time integrator, we use mass-proportional damping to introduce dissipation into the system. The desired force is 
\begin{align}
    \bm{f}_{d} &= - \alpha M \bm{v}
\end{align}
with $\alpha$ the tunable damping coefficient, $M \in \mathbb{R}^{3K \times 3K}$ mass matrix of all element vertices, and $\bm{v} \in \mathbb{R}^{3K}$ all vertex velocities for all $K$ vertices of an element.

\subsection{Contact and Collisions}
Handling contact and collisions accurately and efficiently is a critical aspect of simulating soft body dynamics. Methods such as the active set method~\cite{baraff_large_1998, anitescu_formulating_1997} and the interior point method~\cite{mangoni_primaldual_2018, howell_dojo_2023} have been widely used for solving inequality-constrained-based optimization for contact and collision handling in soft body simulations. However, we opted against the interior point method because of the challenges posed by our non-convex energy formulation. Instead, we employ the SQP method, which belongs to the class of active-set-strategies~\cite{jorge_nocedal_numerical_nodate}.

We formulate collisions as inequality constraints $\bm{h}(\bm{x}) \geq \bm{0}$, where each component of $\bm{h}$ encodes the signed distance between a selected vertex and the contact plane. These constraints are seamlessly integrated into our energy-minimization solver. This method enables stable and efficient resolution of contact events while preserving the structure of the underlying soft body dynamics. For a given vertex position $\bm{x} \in \mathbb{R}^3$ near a contact plane we have
\begin{equation}
h(\bm{x}) \geq \bm{n}^\top (\bm{x} - \bm{p})
\end{equation}
where $\bm{n} \in \mathbb{R}^3$ is the (unit) normal vector of the contact plane and $\bm{p} \in \mathbb{R}^3$ is a point on the plane. 
Let $\mathcal{C}$ be the set of all active contact vertices, then we solve
\begin{align}
\min_{\bm{x}} E(\bm{x}), \quad h_i(\bm{x}) \geq 0 \quad \forall i \in \mathcal{C}.
\end{align}

\section{Results}

To validate the accuracy of our simulator across a range of scenarios, we conduct both rudimentary and sophisticated sim-to-real experiments. System identification (SysId) is performed to determine material parameters that match real-world experimental data. For basic material dynamics validation, we simulate a cantilever and compare its deformation under load with physical experiments, ensuring accurate representation of oscillatory behavior. We further evaluate our framework using objects from the PokeFlex dataset~\cite{obrist_pokeflex_2024} --- a real-world benchmark of volumetric deformable objects through targeted poking by a robotic actuator. These experiments allow us to test contact dynamics and local deformations under external force, further confirming the physical fidelity of our simulations. Additionally, we test the simulator capabilities with a more complex robotic system, a pressure-actuated soft robotic arm~\cite{toshimitsu_sopra_2021}. This test demonstrates the ability of our simulator to model intricate actuation mechanisms and nonlinear deformations accurately, aligning well with observed real-world behavior. We summarize the results of these three sim-to-real experiments in \Cref{table:sim2real}. 

We further detail a use case of our simulation framework for optimization of controller design. The \texttt{Python} interface allows for easy integration with standard optimization libraries. We use the Bayesian optimization provided by Weights and Biases~\cite{biewald_weights_2020}, which allows for easily customizable optimization tasks that can be distributed on multiple computers. 
We optimize muscle activations~\cite{min_softcon_2019} in a simulated soft robotic leg, yielding an efficient strategy for jumping, underscoring the framework strength in enabling optimization of soft robotic systems.

\begin{table}[h]
    \vspace{-0.5em}
    \small
    \centering
    \caption{Parameter domains for each experiment.}
    \label{table:parameter_domains}
    \setlength{\tabcolsep}{3.5pt}
    \renewcommand{\arraystretch}{1.15}
    \begin{tabular}{
        l|
        >{\centering\arraybackslash}m{1.85cm}
        >{\centering\arraybackslash}m{1.85cm}
        >{\centering\arraybackslash}m{1.85cm}
    }
        \toprule
        \textbf{Parameter} \hspace{0.5em} & \textbf{Cantilever} & \textbf{PokeFlex} & \textbf{Soft Arm} \\
        \midrule
        {$\rho$} (\SI{}{\kilo\gram\per\meter^3}) &
        $[700,1300]$ &
        -- &
        $[500,1400]$ \\
        
        $E$ (\SI{}{\kilo\pascal}) &
        $[100, 500]$ &
        $[0.1,100]$ &
        $[150,500]$ \\
        
        $\nu$ (-) &
        $[0.25,0.49]$ &
        $[0.01,0.45]$ &
        $[0.2,0.49]$ \\
        
        $\alpha$ (\SI{}{\per\second}) &
        $[0,30]$ &
        $[1,200]$ &
        -- \\
        \bottomrule
    \end{tabular}
    \vspace{-1em}
\end{table}

\subsection{Cantilever}

\begin{figure}[!b]
    \vspace{-2em}
    \centering
    \includegraphics[width=\columnwidth, trim={0 0.5em 0 0}, clip]{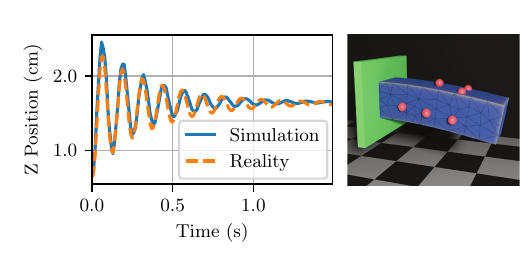}
    \caption{System identification of passive soft cantilever under varying loads. Here we show one trajectory of a \SI{210}{\gram} load, matching phase, frequency, and amplitude of real-world data.}
    \label{fig:sim2real_beam}
\end{figure}

We use the experimental setup from \cite{gao_sim--real_2024}, where $17$ different weights are attached to the tip of a clamped cantilever of size $\SI{10}{\centi\meter} \times \SI{3}{\centi\meter} \times \SI{3}{\centi\meter}$ and suddenly released to observe the natural oscillation of the system, captured in \Cref{fig:sim2real_beam} at $\SI{100}{\hertz}$. 
A Neo-Hookean material is chosen for the cantilever, and the loads are applied as external forces with backward Euler to quickly find a steady-state solution. When releasing the load, we use Crank-Nicolson timestepping to preserve energy during oscillation and manually introduce dissipation through mass-proportional damping. For the SysId we search over mass density $\rho$, Young's modulus $E$, Poisson's ratio $\nu$ and damping parameter $\alpha$, see \Cref{table:parameter_domains}.

We compute the error by taking the average distance norm of the $N$ motion markers over all $R$ weights and $T$ timesteps:
\begin{equation}
    \mathcal{E} := \frac{1}{R \cdot T \cdot N}\sum_{j=1}^R\sum_{t=1}^T\sum_{i=1}^N \left\| \mathbf{x}^j_{t,i}-\overline{\mathbf{x}}^j_{t,i} \right\|
\label{eq:marker_errors}
\end{equation}
where $\mathbf{x}^j_{t,i}$ denotes the $i$-th marker position at the $j$-th trajectory and time step $t$, and $\overline{\mathbf{x}}^j_{t,i}$ is the experimental ground truth. We find an optimized error of \SI{4.98}{\milli\meter} with a density of \SI{1210}{\kilo\gram\per\meter^3}, Young's modulus of \SI{234.9}{\kilo\pascal}, Poisson's ratio of \SI{0.439}{}, and a mass damping coefficient $\alpha$ of \SI{9.11}{\per\second}. \Cref{fig:overview} shows the simulated cantilever deformation next to the real-world experiment. The simulator accurately reproduces the static tip deflection and dynamic oscillation frequency after release as seen in \Cref{fig:sim2real_beam}, confirming the physical fidelity of the identified values, which is close to material parameters found in related works~\cite{rothemund_soft_2018}. The close alignment of marker trajectories in both amplitude and phase demonstrates that our simulator captures both the elastic response and energy dissipation behavior observed experimentally. 


\subsection{PokeFlex}
\label{sec:poke}

\begin{figure}[!t]
    \centering
    \includegraphics[width=\linewidth]{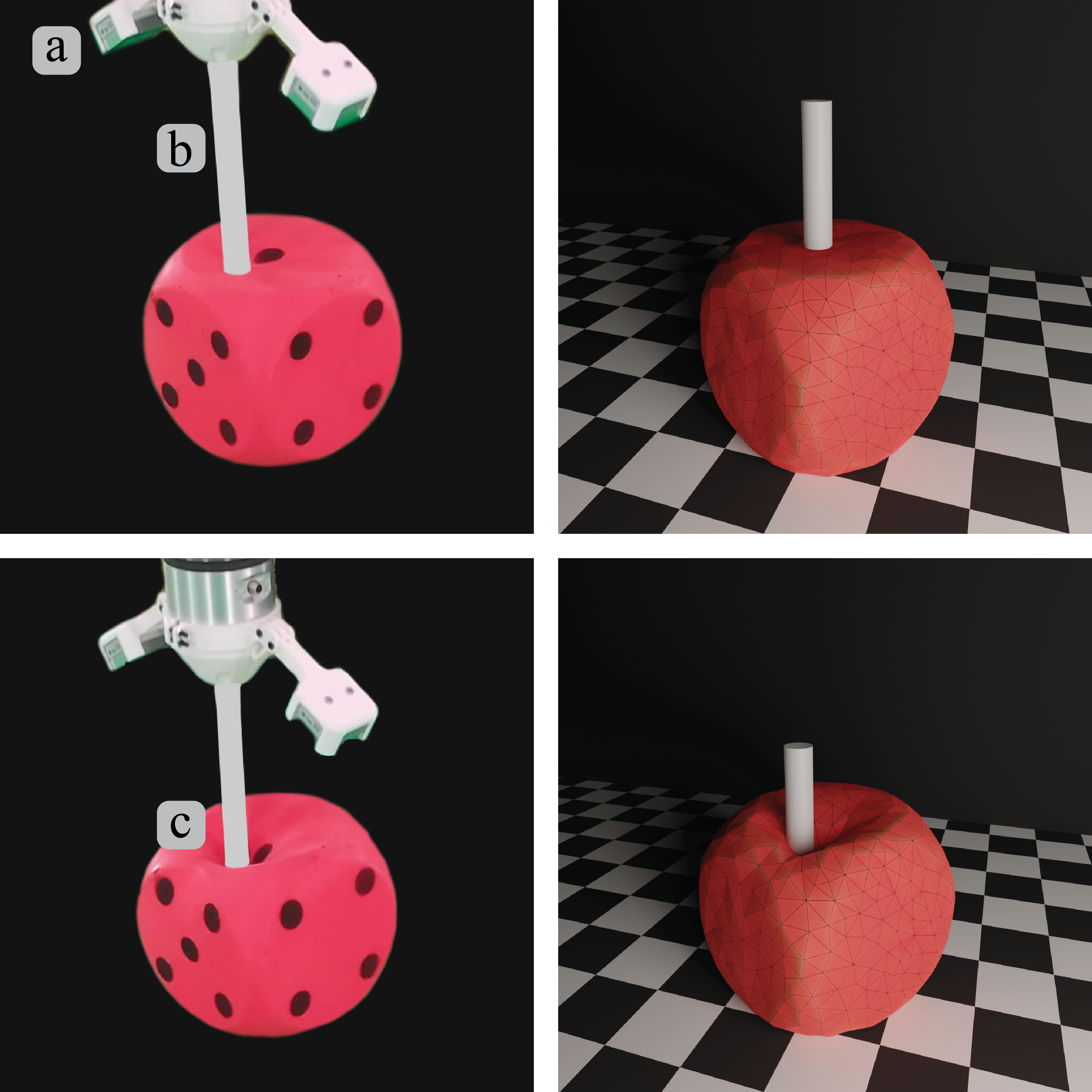}
    \caption{a) End-effector, b) contact cylinder (colored in image as the rod was transparent), and c) indentation area. The sim-to-real comparison demonstrates the ability of our simulator to accurately reproduce contact-induced deformations and overall shape fidelity of highly nonlinear materials.}
    \label{fig:sim2real_pokeflex}
    \vspace{-1em}
\end{figure}

We validated the ability of our simulator to model contact dynamics and highly nonlinear deformations under external forces using the PokeFlex dataset~\cite{obrist_pokeflex_2024}, which provides real-world multimodal data of deformable objects captured with a 360$^\circ$ volumetric system, including reconstructed 3D meshes and contact interactions generated by a robotic arm with a cylindrical end-effector. In our simulation, we replicated the poking interaction on a $\SI{16}{cm}$-sided soft cube, using identical actuator trajectories to those recorded in the dataset. The soft cube was modeled with a stable Neo-Hookean elasticity from \cite{smith_stable_2018} and mass-proportional damping, and the time integration was performed using Crank-Nicolson scheme. Using the same SysId method as in the previous subsection, we optimize over Young's modulus $E$, Poisson's ratio $\nu$ and mass-damping parameter $\alpha$, refer to \Cref{table:parameter_domains}. We evaluated the fit between simulation and experiment using the mean Chamfer distance across all timesteps $T$ and trajectories $R$, defined for two point clouds as 
\begin{align}
\mathcal{E}
&:= \frac{1}{R \cdot T}\sum_{j=1}^{R}\sum_{t=1}^{T} \mathrm{CD}\!\left(P_{j,t},\,Q_{j,t}\right)^{\frac{1}{2}} \label{eq:mean_cd}\\[0.3em]
\mathrm{CD}(P,Q)
&= \frac{1}{|P|}\sum_{p\in P}\min_{q\in Q}\|p-q\|^2
 + \frac{1}{|Q|}\sum_{q\in Q}\min_{p\in P}\|q-p\|^2 \nonumber
\end{align}
where $P_{j,t}$ and $\,Q_{j,t}$ are the simulated and experimental surface point clouds of trajectory $j$ at timestep $t$, respectively. The Chamfer distance measures the geometric discrepancy between two point clouds by computing, for each point in one cloud, the distance to its nearest neighbor in the other. The bidirectional averaging ensures that both surface regions contribute to the final error, where smaller values indicate closer alignment between simulation and experiment. The optimized parameters resulting in the lowest mean Chamfer distance were found to be $E = \SI{1346}{Pa}$, $\nu = 0.018$ and $\alpha = \SI{170}{\per\second}$. We find a mean Chamfer distance of \SI{6.9}{mm}, which is small relative to the cube edge length \SI{160}{mm} and indicates strong geometric correspondence between simulation and experiment, as seen in \Cref{fig:sim2real_pokeflex}. 

\begin{figure}[!t]
    \centering
    \includegraphics[width=\columnwidth, trim={0 0.5em 0 0}, clip]{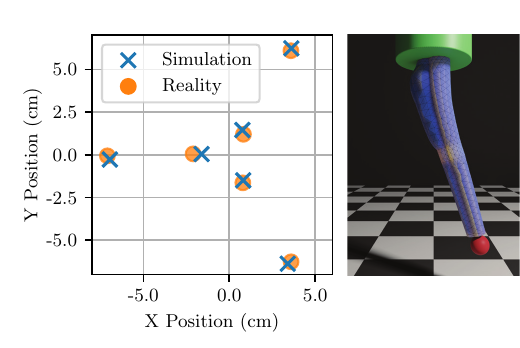}
    \caption{System identification of the actuated pneumatic soft arm for six quasistatic tip positions, where each of the six chambers is individually actuated at a pressure of \SI{60}{\kilo\pascal}.}
    \label{fig:sim2real_sopra}
    \vspace{-1em}
\end{figure}

\subsection{Soft Arm}

Next, our simulator is applied on actuated soft robots such as a pneumatically-actuated soft arm \cite{toshimitsu_sopra_2021}, approximately \SI{30}{\centi\meter} long. We model the six pneumatic chambers with uniform pressure forces per chamber. External fiber reinforcements are neglected and we use a homogeneous Neo-Hookean material for the whole arm structure. For the SysId we search over mass density $\rho$, Young's modulus $E$ and Poisson's ratio $\nu$, see \Cref{table:parameter_domains}. We use the same marker error defined in \Cref{eq:marker_errors}, where this time we have 6 steady states of the robot when each chamber is inflated at $\SI{60}{\kilo\pascal}$. A visual comparison can be seen in \Cref{fig:overview}. The optimized error is \SI{2.53}{\milli\meter} with a density of \SI{1167}{\kilo\gram\per\meter^3}, Young's modulus of \SI{346.5}{\kilo\pascal}, and a Poisson's ratio of \SI{0.358}{}. We visualize the individual errors for all six pressurized chambers in \Cref{fig:sim2real_sopra}. We observe that even without modeling the fiber-reinforcements we can find material parameters that allow for matching tip behavior.

\begin{table}[h]
    \vspace{-0.5em}
    \small
    \centering
    \caption{Number of degrees of freedom, trajectories sample size, and mean error (average distance norm for cantilever and soft arm, Chamfer distance for PokeFlex).}
    \label{table:sim2real}
    \setlength{\tabcolsep}{3.5pt}
    \renewcommand{\arraystretch}{1.15}
    \begin{tabular}{
        l
        >{\centering\arraybackslash}m{1.85cm}
        >{\centering\arraybackslash}m{1.85cm}
        >{\centering\arraybackslash}m{1.85cm}
    }
        \toprule
         & \textbf{Cantilever} & \textbf{PokeFlex} & \textbf{Soft Arm} \\
        \midrule
        \textbf{nDoF} & 564  & [3813, 5184] & 3249 \\
        \textbf{Trajectories} & 17 & 10 & 6 \\
        \textbf{Error} (\SI{}{\milli\meter}) & $4.98 \pm 1.32$ & $6.94 \pm 1.66$ & $2.53 \pm 1.16$ \\
        \bottomrule
    \end{tabular}
    \vspace{-1em}
\end{table}


\subsection{Muscle-Actuated Leg}

\begin{figure}[!b]
    \vspace{-1em}
    \centering
    \includegraphics[width=\columnwidth, trim={0 0.5em 0 0}, clip]{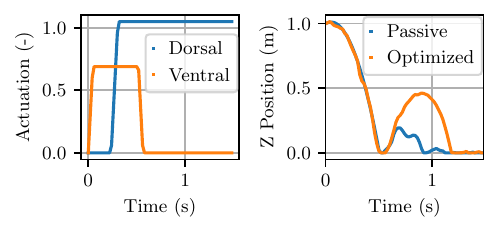}
    \caption{(Left) Actuation signal of ventral and dorsal contraction over time. (Right) The trajectory of the lowest vertex of the unactuated and optimized actuated leg is shown, reaching a maximum optimized jumping height of \SI{0.463}{\meter}.}
    \label{fig:leg_signal}
\end{figure}

Lastly, we showcase that our simulation can be used in a more custom optimization setting. Integration with a distributed Bayesian optimization framework such as Weights and Biases allows optimization of arbitrary user-defined objectives. We choose a muscle-actuated leg robot that we require to jump as high as possible after colliding with the ground. This example also showcases hexahedral meshing, where we define the robot on a voxel grid, with a total height of \SI{0.5}{m}. The leg has two groups of antagonistic muscles, 10 voxels in the back (dorsal) and 10 voxels in the front (ventral) that can contract up to half their length along the z-axis, shown in \Cref{fig:leg_signal}.

The actuation of both the ventral and dorsal muscle group are step functions, where we define the magnitude of contraction and when to contract/relax. The whole leg jump happens in the span of \SI{1.5}{\second}, where an initial forward velocity is given of \SI{0.5}{\meter\per\second}. The ventral muscle starts contracted with contraction strength $a_v$ and relaxes at a given time $t_v$, while the dorsal muscle starts contracting with strength $a_d$ at a certain time $t_d$, which can be seen in \Cref{fig:leg_signal}. The actuation signal parametrized by four variables is shown in \Cref{tab:leg_domain}.

\begin{table}[h]
    \vspace{-0.5em}
    \small
    \centering
    \caption{Parameters for jumping leg optimization.}
    \label{tab:leg_domain}
    \setlength{\tabcolsep}{3.5pt}
    \renewcommand{\arraystretch}{1.15}
    \begin{tabular}{
        l|
        >{\centering\arraybackslash}m{1.2cm}
        >{\centering\arraybackslash}m{1.2cm}
        >{\centering\arraybackslash}m{1.2cm}
        >{\centering\arraybackslash}m{1.2cm}
    }
        \toprule
        \textbf{Leg Parameter} \hspace{0.5em} & {$a_{v}$} (-) & {$a_{d}$} (-) & {$t_{v}$} (\SI{}{\second}) & {$t_{d}$} (\SI{}{\second}) \\
        \midrule
        Search Range &
        $[0.0, 2.0]$ &
        $[0.0, 2.0]$ &
        $[0.1, 0.9]$ &
        $[0.1, 0.9]$ \\
        
        Optimal Value &
        $0.69$ &
        $1.05$ &
        $0.54$ &
        $0.27$ \\
        \bottomrule
    \end{tabular}
    \vspace{-0.5em}
\end{table}

\begin{figure}[!t]
    \centering
    \includegraphics[width=\linewidth]{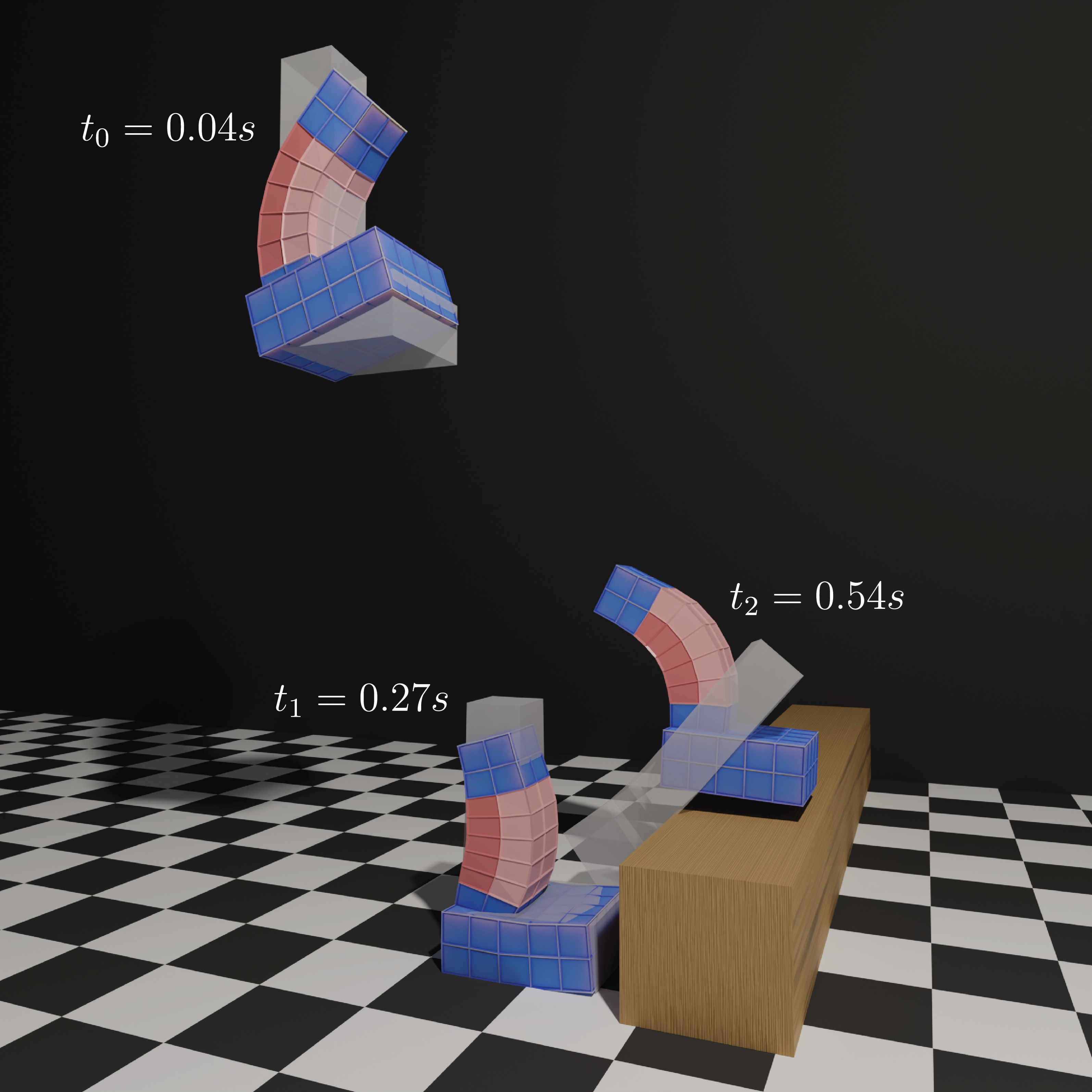}
    \caption{Jumping performance of the simulated muscle-actuated leg. The figure shows three key timesteps for the passive and optimized actuation sequence. Only the optimized leg successfully clears the obstacle, the passive leg fails to do so and collides. 
    }
    \vspace{-1em}
    \label{fig:leg}
\end{figure}

We show in \Cref{fig:leg} how an unactuated leg cannot reach the height of an obstacle, while with the optimized actuation signal for contraction of both muscle groups we can jump a far larger height. The optimized variables we find are \SI{0.69}{} for the starting contraction of the front muscle, and \SI{1.05}{} for back muscle contraction at the end, where the front muscle relaxes at \SI{0.54}{\second} and the back muscle contracts at \SI{0.27}{\second}. The maximal height reached is \SI{0.463}{\meter}.

\section{Conclusion}

In this work, we presented SORS, a modular, high-fidelity simulation framework for soft robotic systems. Built upon constrained energy minimization, our simulator unifies nonlinear elasticity, actuation, contact, and more through three extensible interfaces: energies, forces, and constraints. In the rapidly evolving field of soft robotics, the provided modular and customizable structure is crucial for adapting to new physical models and actuation mechanisms. Through extensive sim-to-real experiments, including cantilever beam deformation, contact interactions from the PokeFlex dataset, and a pneumatic soft arm, we validated the framework’s ability to reproduce real-world behavior with millimeter-scale accuracy. Furthermore, we demonstrated its use by optimizing muscle activation signals for a soft robotic leg, highlighting integration with optimization pipelines. Future work will focus on GPU acceleration for large-scale simulations and the incorporation of frictional and multi-body contact models. 
By combining physical accuracy, extensibility, and open accessibility, SORS establishes a new foundation for advancing the design, control, and analysis of next-generation soft robots.

\printbibliography
\end{document}